\documentclass{article} 
\usepackage{iclr2026_conference,times}
\iclrfinalcopy

\usepackage{hyperref}
\usepackage{url}
\usepackage[utf8]{inputenc} 
\usepackage[T1]{fontenc}    
\usepackage{hyperref}       
\usepackage{url}            
\usepackage{booktabs}       
\usepackage{amsfonts}       
\usepackage{nicefrac}       
\usepackage{microtype}      
\usepackage{xcolor}         
\usepackage{graphicx}
\usepackage{subfigure}
\usepackage{mathtools}
\usepackage{amsthm}
\usepackage{amsmath}
\usepackage{amssymb}
\usepackage{extarrows}
\usepackage{diagbox}
\usepackage{microtype}
\usepackage{multirow}
\usepackage[most]{tcolorbox}
\usepackage{makecell}

\newcommand{\method}{\texttt{EBM-CoT}}

\usepackage{amsmath,amsfonts,bm}

\def\eqref#1{equation~\ref{#1}}

\def\1{\bm{1}}

\DeclareMathAlphabet{\mathsfit}{\encodingdefault}{\sfdefault}{m}{sl}
\SetMathAlphabet{\mathsfit}{bold}{\encodingdefault}{\sfdefault}{bx}{n}

\definecolor{text}{rgb}{0.4,0.1,0.4}

\usepackage{algorithm}
\usepackage{algorithmic}

\title{Think Consistently, Reason Efficiently: Energy-Based Calibration for Implicit Chain-of-Thought}

\author{
Zhikang Chen\textsuperscript{1}\quad 
Sen Cui\textsuperscript{2}\quad 
Deheng Ye\textsuperscript{3}\quad 
Yu Zhang\textsuperscript{~4}\quad 
Yatao Bian\textsuperscript{\dag~5}\quad 
Tingting Zhu\textsuperscript{\dag~1}\\
\textsuperscript{1}University of Oxford\quad
\textsuperscript{2}Tsinghua University\quad
\textsuperscript{3}Tencent\\
\textsuperscript{4}Southern University of Science and Technology\quad
\textsuperscript{5}National University of Singapore\\[3pt]
\textsuperscript{\dag}Corresponding authors
}

\makeatletter
\renewcommand{\headrulewidth}{0pt}  
\makeatother
\begin{document}

\maketitle

\begin{abstract}
Large Language Models (LLMs) have demonstrated strong reasoning capabilities through \emph{Chain-of-Thought} (CoT) prompting, which enables step-by-step intermediate reasoning. 
However, explicit CoT methods rely on discrete token-level reasoning processes that are prone to error propagation and limited by vocabulary expressiveness, often resulting in rigid and inconsistent reasoning trajectories. 
Recent research has explored implicit or continuous reasoning in latent spaces, allowing models to perform internal reasoning before generating explicit output. 
Although such approaches alleviate some limitations of discrete CoT, they generally lack explicit mechanisms to enforce consistency among reasoning steps, leading to divergent reasoning paths and unstable outcomes. 
To address this issue, we propose \method{}, an Energy-Based Chain-of-Thought Calibration framework that refines latent thought representations through an energy-based model (EBM). 
Our method dynamically adjusts latent reasoning trajectories toward lower-energy, high-consistency regions in the embedding space, improving both reasoning accuracy and consistency without modifying the base language model. 
Extensive experiments across mathematical, commonsense, and symbolic reasoning benchmarks demonstrate that the proposed framework significantly enhances the consistency and efficiency of multi-step reasoning in LLMs.

\end{abstract}

\section{Introduction}


In recent years, Large Language Models (LLMs) have achieved groundbreaking progress in natural language processing \citep{achiam2023gpt, dubey2024llama, yang2025qwen3, guo2025deepseek, team2025kimi}, demonstrating not only remarkable language understanding and generation capabilities, but also strong performance in reasoning tasks. Chain-of-Thought (CoT) has emerged as a central mechanism for enhancing the reasoning capabilities of large language models (LLMs), enabling them to generate intermediate reasoning steps that progressively lead to the final answer. \cite{wei2022chain} first demonstrated that prompting models to generate intermediate reasoning steps can substantially improve performance in arithmetic and commonsense reasoning tasks, revealing the emergence of step-by-step reasoning abilities as the model scale increases. This study established CoT as a mechanism that not only guides explicit reasoning but also reflects the model's intrinsic thought process. However, traditional explicit CoT methods rely on discrete, token-level reasoning processes, which are prone to error propagation during generation and constrained by the limited expressiveness of the vocabulary space -- resulting in rigid reasoning paths and poor robustness. 
To address these issues, recent research has explored \emph{implicit reasoning} processes, allowing models to ``think'' in a continuous representation space before producing the final output. For example, methods such as Coconut \citep{hao2024training} and CCoT \citep{cheng2024compressed} compress discrete reasoning sequences into continuous latent variables, enabling efficient \emph{soft reasoning}. Despite their promise, these approaches rely on full-model fine-tuning, which is computationally expensive and often leads to \emph{catastrophic forgetting} \citep{mccloskey1989catastrophic, fernando2024mitigating, wu2025mitigating} -- a phenomenon in which the model loses previously acquired capabilities after learning new continuous reasoning representations, ultimately degrading reasoning performance. SoftCoT \citep{xu2025softcot} mitigates this issue by freezing the base and assistant models and introducing a projection model that generates soft thought tokens, which are projected into the embedding space of the base model by linear mapping. This design effectively alleviates catastrophic forgetting. However, SoftCoT still suffers from a fundamental limitation: since it lacks explicit consistency constraints during training, the generated reasoning trajectories tend to diverge at inference time, resulting in inconsistent outcomes when multiple CoT are sampled. Consequently, multiple CoT samples are typically required to be generated and ranked to achieve satisfactory performance, at the cost of significantly reduced efficiency.

\begin{figure}[t!]
  \setlength{\abovecaptionskip}{-0.1cm}
  \setlength{\belowcaptionskip}{-0.2cm}
  \centering

    \includegraphics[width=\columnwidth]{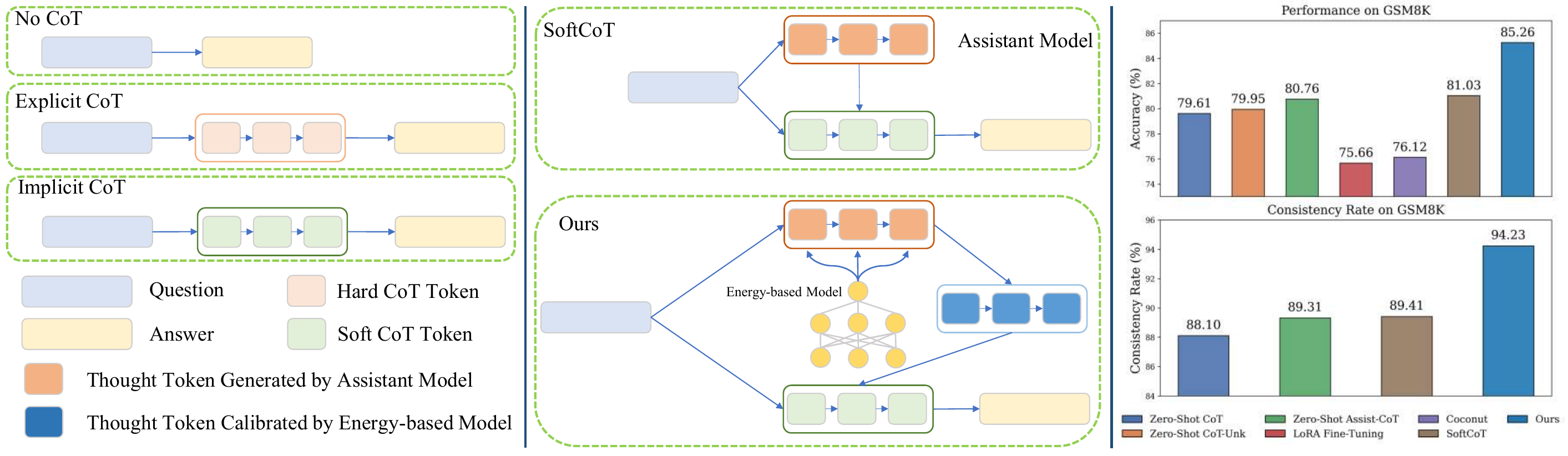}
  \caption{
    Comparison of reasoning paradigms and the proposed energy-based calibration framework.
    \textbf{Left Figure:} Different forms of reasoning in language models, including No CoT, Explicit CoT, and Implicit CoT.
    \textbf{Middle Figure:} Our method integrates an Energy-Based Model (EBM) to calibrate the latent thought tokens generated by the assistant model, producing refined soft thoughts with improved coherence and consistency.
    \textbf{Right Figure:} Experimental results on GSM8K show that our approach, implemented on top of LLaMA-3.1-8B-Instruct, achieves superior accuracy and substantially higher consistency compared to previous CoT variants, effectively matching the performance of multi-CoT reasoning with a single CoT using calibrated thought tokens. 
    }

  \label{fig:overview}
  \vspace{-0.2cm}
\end{figure}

To address the aforementioned limitations, we propose an \emph{implicit Chain-of-Thought calibration framework} based on an \emph{Energy-Based Model (EBM)}, called \method{}. An overview of the proposed framework is illustrated in Figure~\ref{fig:overview}.
The left part of the figure contrasts different reasoning paradigms, including No CoT, Explicit CoT, and Implicit CoT. Unlike prior approaches that rely solely on continuous representation mapping, \method{} introduces a differentiable energy function $E_{\phi}(x)$ defined in the thought embedding space, as illustrated in the middle part of Figure~\ref{fig:overview}. This energy function explicitly calibrates the latent thought tokens, guiding them toward lower-energy regions that correspond to more task-relevant and logically consistent reasoning trajectories. Although our model performs energy-based calibration only at the embedding level, the learnt energy landscape implicitly integrates multiple potential reasoning directions in the continuous space. In our framework, each soft thought token $l_t$ represents an implicit reasoning step within the chain-of-thought process. Although referred to as a ``token'' for its sequential and compositional role, it is realised as a continuous soft thought embedding in the latent space. 
This design preserves the token-level reasoning structure while enabling differentiable operations such as gradient-based energy optimisation and Langevin dynamics sampling. As a result, calibrating the soft thought tokens enables a single chain of thought to approximate the reasoning performance of multi-CoT ensembles. During training, we perform multi-step energy-guided updates in the thought embedding space using Langevin dynamics. The energy gradient $-\nabla_x E_{\phi}(x)$ provides a corrective direction towards lower-energy regions (higher-consistency), automatically refining the model’s reasoning trajectory at each inference step. Moreover, at the outer level of optimisation, we jointly train the language modelling objective with the EBM objective, comprising both an energy ranking constraint and a consistency constraint, to align the direction of the gradient of the energy model with that of the language model. Consequently, the energy function not only learns \emph{what constitutes a plausible reasoning representation}, but also \emph{how to proactively refine implausible reasoning states} during the reasoning process. At inference time, \method{} leverages the learnt energy function to perform a few steps of Langevin calibration in the thought embedding space, guiding the reasoning state from higher-energy (low-consistency) regions toward lower-energy (high-consistency) ones. The calibrated embeddings are then fed into the frozen base model for subsequent generation, thereby achieving dynamic reasoning calibration and enhanced robustness without any parameter modification to the base and assistant LLM. The right part of Figure~\ref{fig:overview} presents the empirical impact of our calibration mechanism, specifically, the consistency rate is defined as
$\text{Consistency Rate} = \frac{\text{Acc(pass@$1$)}}{\text{Acc(pass@$N$)}} \times 100\%,$
which measures how consistently the model reaches correct conclusions when multiple reasoning chains (pass@$N$) are sampled for the same input. 
As shown in this figure, we set $N=10$ following the standard self-consistency protocol. 
A higher consistency rate indicates a more stable and reliable reasoning process, 
and our energy-based calibration significantly enhances both accuracy and consistency across sampled reasoning paths.
These results indicate that aligning the latent thought tokens through energy-based calibration allows a single CoT to achieve performance comparable to that of multi-CoT ensembles ($N=10$). 
Our key contributions are summarized as follows:

\begin{enumerate}
\item We introduce a novel perspective on reasoning in large language models by employing an Energy-Based Model (EBM) to enhance Chain-of-Thought (CoT) reasoning.
    Unlike standard auto-regressive models that ensure only local token-level consistency, our approach explicitly promotes \emph{global consistency} and \emph{logical coherence} across the entire latent reasoning trajectory.
    This establishes a new bridge between energy-based modeling and implicit reasoning in language models.
\item We design a hybrid, multi-stage architecture that separates implicit \emph{thinking} in a latent space from explicit textual reasoning and answer generation.
    By applying Langevin dynamics to continuous latent embeddings, the EBM performs energy-based calibration of internal thoughts before text generation occurs,
    refining latent reasoning states toward low-energy, high-consistency regions in the embedding space.
\item Extensive experiments on mathematical, commonsense, and symbolic reasoning benchmarks demonstrate that our method significantly improves both reasoning accuracy and consistency.
    Notably, the model achieves strong single-chain performance (N=1), narrowing the gap with multi-chain self-consistency results while maintaining substantially higher efficiency.
\end{enumerate}

\section{Related Work}


\textbf{CoT Performance Enhancement}. \citet{wang2022self} introduced the Self-Consistency decoding strategy, which strengthens robustness by sampling multiple reasoning paths and selecting the most consistent final answer. Tree-of-Thoughts (ToT) \citep{yao2023tree} extends the linear Chain-of-Thought paradigm into a tree-structured reasoning framework, enabling models to explore, evaluate, and backtrack among multiple reasoning paths. Graph-of-Thoughts (GoT) \citep{besta2024graph} further generalizes this idea by modelling reasoning nodes as a graph structure, supporting multi-path integration and feedback between thoughts. To make such a global search more efficient, Chain-of-Preference Optimization (CPO) \citep{zhang2024chain} uses implicit preference data embedded in ToT search trees to fine-tune models, allowing CoT to achieve reasoning quality close to ToT at a lower inference cost. \citet{jiang2025learning} introduces the Energy Outcome Reward Model (EORM), an energy-based reranker that scores and orders Chain-of-Thought candidates by outcome quality. However, such approaches tend to significantly reduce the efficiency of CoT-based inference. 

\textbf{CoT Reasoning Efficiency Improvement}. To improve the efficiency of CoT-based reasoning, researchers have proposed various implicit or compressed forms of CoT. \cite{cheng2024compressed} introduced Compressed Chain-of-Thought (CCoT), which compresses explicit linguistic reasoning into continuous latent “contemplation tokens,” substantially improving inference efficiency. SoftCoT \citep{xu2025softcot} further advanced this idea by performing soft chain-of-thought reasoning in the continuous latent space: an auxiliary model generates soft thought tokens that are mapped into the embedding space of the frozen main model, thus improving reasoning efficiency while mitigating catastrophic forgetting. \cite{shen2025codi} further proposed the CODI framework, which transfers the reasoning ability of explicit CoT into the continuous latent space through self-distillation. \citet{xia2025tokenskip} proposed the TokenSkip framework, which enables controllable Chain-of-Thought (CoT) compression by skipping semantically redundant reasoning tokens. 
TokenSkip learns to identify and remove low-importance tokens based on token-level importance estimation, thereby reducing CoT length while maintaining reasoning accuracy. Collectively, these methods signify an evolution of reasoning from explicit linguistic chains toward internal continuous representations. 





\textbf{Energy-based Model}. In recent years, Energy-Based Models (EBMs) \citep{hopfield1982neural, hinton1983optimal, ren2024carbonnovo, wang2025energy, balcerak2025energy, zhang2025right} have emerged as a unified probabilistic modelling paradigm, demonstrating remarkable expressiveness across generative modelling, model alignment, and decision optimisation. Unlike approaches that explicitly model probability distributions, EBMs define distributions implicitly through an energy function, offering strong compositionality, constraint flexibility, and compatibility with other model architectures. In the optimisation process, \citep{pang2024arm} and \citep{hong2024energy} proposed ARM and EPA, respectively, which replace RLHF with residual energy models and perform preference optimisation through forward KL divergence minimisation, achieving both stability and efficiency.  \cite{zhou2024antigen} further introduced ABDPO, an energy-based direct preference optimisation framework for antibody design, where residue-level energy decomposition guides diffusion models to generate structurally plausible and high-affinity antibodies. In generative modelling, \cite{xu2024energy} proposed the Energy-based Diffusion Language Model, which integrates energy functions with discrete diffusion processes to mitigate parallel decoding errors and improve text generation quality. \cite{zhang2025energymogen} introduced EnergyMoGen, which combines latent-space and semantic energy to enable compositional generation of multi-concept motions. \cite{wang2025equilibrium} proposed Equilibrium Matching, which unifies flow and diffusion models under a balanced energy field, supporting efficient sampling via gradient-based optimizzation. Overall, these studies highlight the broad applicability of EBMs across various domains.


\section{Preliminaries}

In this section, we formalize the reasoning process of large language models (LLMs) and introduce the foundational concepts underlying our method.
Section~\ref{sec:problem-definition} defines the problem setup and decomposes the reasoning process into three auto-regressive stages: Thinking, Reasoning, and Answer Generation.
Section~\ref{sec:ebm} then introduces the principles of energy-based modeling, which serve as the theoretical basis for our subsequent energy-calibrated reasoning framework.

\subsection{Problem Definition}
\label{sec:problem-definition}

We decompose the overall reasoning process of a large language model (LLM)
into three auto-regressive stages that are consistent with the
energy-adjusted chain-of-thought formulation introduced above.
Let the input be $x$, and let the model generate a sequence of intermediate
representations or reasoning steps $y_{1:T} = (y_1, y_2, \dots, y_T)$
with contextual states $c_t = (x, y_{<t})$.

Large language models (LLMs) often generate intermediate reasoning steps,
or \emph{chains of thought} (CoT), to solve complex reasoning tasks.
Given an input $x$, the model produces a reasoning trajectory
\[
    y_{1:T} = (y_1, y_2, \dots, y_T),
\]
where each token $y_t$ corresponds to a partial reasoning step conditioned on the context
$c_t = (x, y_{<t})$.
Under the standard auto-regressive factorization, 
the conditional distribution over the entire reasoning sequence is
\[
    p(y_{1:T}\mid x) = \prod_{t=1}^T p(y_t \mid c_t),
    \qquad c_t = (x, y_{<t}).
\]
In conventional language models, each conditional probability $p(y_t \mid c_t)$ 
is parameterized by a softmax over the model logits. The probability of selecting a specific token \(v\) from the vocabulary \(\mathcal{V}\) as the next token \(y_{t}\) is calculated as:
\[
    p(y_t = v \mid c_t)
    = \frac{\exp(W_v^\top h_t + b_v)}
           {\sum_{v' \in \mathcal{V}} \exp(W_{v'}^\top h_t + b_{v'})},
\]
where \(h_{t}=f_{\theta }(c_{t})\) is the hidden representation of the context \(c_{t}\), \(f_{\theta }\) is the function of the neural network, and \(\mathcal{V}\) is the vocabulary.
This formulation can be equivalently viewed as a local energy-based model,
where each next-token prediction defines a locally normalized energy landscape. The energy for a specific prediction $y_{t}$ given the context \(c_{t}\) is
\[
    E_\theta(c_t, y_t) = - (W_{y_t}^\top h_t + b_{y_t}),
\]
and the conditional probability of the next token $y_{t}$ can be expressed using a Boltzmann distribution
\[
    p_\theta(y_t \mid c_t)
    = \frac{\exp[-E_\theta(c_t, y_t)]}
           {\sum_{y' \in \mathcal{V}} \exp[-E_\theta(c_t, y')]}.
\]
Hence, next-token prediction in auto-regressive language modeling 
can be interpreted as a sequence of locally normalized EBMs defined over discrete vocabulary tokens.
However, this formulation captures only token-level dependencies and provides local normalization at each step,
without enforcing global consistency across reasoning trajectories.
In the following sections, we extend this energy-based perspective from discrete token prediction
to a continuous latent space, where reasoning is modeled as an energy-adjusted process over latent thought embeddings.

\paragraph{Thinking Stage}
The assistant model first produces a sequence of latent thought embeddings
$L = \{l_1, \dots, l_n\}$ that represent continuous reasoning states and capture the model’s internal thought process:
\begin{equation}
    p_a(L \mid x)
    = \prod_{i=1}^{n} p_a(l_i \mid c_i),
    \qquad c_i = (x, l_{<i}),
\end{equation}
where $p_a(l_i \mid c_i)$ denotes the conditional distribution of the assistant model over latent thoughts
given the input $x$ and the previously generated latent sequence $l_{<i}$.
Each latent thought $l_i$ is a continuous embedding derived from the hidden representations of the assistant model.
These latent embeddings are then projected into the base model’s input space before being refined by the Energy-Based Model (EBM) in the subsequent calibration step.

\paragraph{Reasoning Stage}
Conditioned on latent thoughts $L$ generated by the assistant model, the base model generates an explicit
textual reasoning trajectory $R = \{r_1, \dots, r_m\}$:
\begin{equation}
    p_\theta(R \mid x, L)
    = \prod_{j=1}^{m} p_\theta(r_j \mid c_j),
    \qquad c_j = (x, L, r_{<j}),
\end{equation}
where $p_\theta(r_j \mid c_j)$ represents the conditional probability of the base model of producing
the $j$-th reasoning token based on input $x$, latent thoughts $L$,
and previously generated reasoning tokens $r_{<j}$.

\paragraph{Answer Generation Stage}
Finally, the base model synthesizes the reasoning trace into an answer
sequence $A = \{a_1, \dots, a_k\}$:
\begin{equation}
    p_\theta(A \mid x, L, R)
    = \prod_{k=1}^{o} p_\theta(a_k \mid c_k),
    \qquad c_k = (x, L, R, a_{<k}).
\end{equation}

\paragraph{Overall Formulation}
The complete conditional distribution of the reasoning process can thus be written as
\begin{equation}
    p(L, R, A \mid x)
    = p_a(L \mid x)\;
      p_\theta(R \mid x, L)\;
      p_\theta(A \mid x, L, R),
\end{equation}
which hierarchically decomposes the reasoning process into latent thinking, explicit reasoning, and answer generation stages. 
This hierarchical decomposition shows that the Thinking, Reasoning, and Answer stages are not only conditioned on input $x$, but also closely depend on the latent thought tokens $L$ that mediate intermediate reasoning.
When the input $x$ is fixed, improving the consistency of reasoning therefore reduces to refining these latent thought tokens, which serve as key carriers of the reasoning dynamics. Since our method performs calibration on the latent thought tokens, in the following derivations, we instantiate the generic variable $y$ as the latent thought embedding $l$.

\subsection{Energy-based Models for CoT}
\label{sec:ebm}

Energy-based models (EBMs) define a scalar energy function
$E_\phi(l)$ that assigns low energy to preferred configurations and
high energy to unlikely ones.
The corresponding probability density $p_\phi(l)$ follows the Boltzmann distribution:

\[
    p_\phi(l)
    = \frac{\exp[-E_\phi(l)]}{Z_\phi},
    \qquad
    Z_\phi = \int \exp[-E_\phi(l)]\,dl.
\]

In the implicit CoT setting, the energy-based distribution can be expressed as
\[
    p_\phi(l \mid c)
    \propto
    p_a(l \mid c)\,
    \exp\!\left(-E_\phi(c, l)/{T}\right),
\]
where $T$ is a temperature
parameter that controls the strength of the energy scaling.
The energy term reweights the base distribution, enabling the model
to prefer output with lower energy -- more coherent or
semantically consistent responses.

As derived in Section \ref{sec:problem-definition}, the next-token prediction can be viewed as a local energy-based model,
each capturing short-range consistency but not explicitly optimized for global reasoning coherence. 
Conventional language modeling maximizes the log-likelihood of each next token,
where the probability \(p(l_{t}\mid c_{t})\) is normalized independently at each step.
However, this local normalization does not explicitly encourage global consistency or logical coherence across the entire reasoning sequence.  
To address this limitation, we reinterpret each conditional distribution
$p(l_t \mid c_t)$ as an energy-calibrated distribution
$\tilde{p}_\phi(l_t \mid c_t)$,
where the energy function $E_\phi(c_t, l_t)$ modulates the likelihood of the base model according to a learned notion of the quality of reasoning.
This formulation provides a unified view of auto-regressive language modelling within the energy-based modelling framework.

Sampling from EBMs typically relies on algorithms like Markov Chain Monte Carlo. For EBMs defined over continuous spaces, unadjusted Langevin dynamics are used, which iteratively refines a sample \(y^{(s)}\) according to
\[
    l^{(s+1)} = l^{(s)}
      - \eta \nabla_l E_\phi(c, l^{(s)})
      + \sqrt{2\eta}\,\varepsilon^{(s)},
      \qquad
      \varepsilon^{(s)} \sim \mathcal{N}(0, I),
\]
forming a Markov chain that converges to the stationary distribution
$p_\phi(l \mid c) \propto \exp[-E_\phi(c, l)]$.

\section{Methodology}

In this section, we introduce our proposed method.
Section~\ref{41} presents the overall framework of EBM-CoT, which incorporates energy-based calibration into the Chain-of-Thought reasoning process.
Section~\ref{42} details the design of the loss function used to jointly optimize language modeling and energy-based objectives.
Section~\ref{43} describes the optimization process, including Langevin refinement and training strategies. Additional proofs and algorithmic details are provided in Appendices \ref{theory} and \ref{alg}.

\subsection{EBM-CoT}
\label{41}

\begin{figure}[t!]
  \setlength{\abovecaptionskip}{-0.1cm}
  \setlength{\belowcaptionskip}{-0.2cm}
  \centering

    \includegraphics[width=\columnwidth]{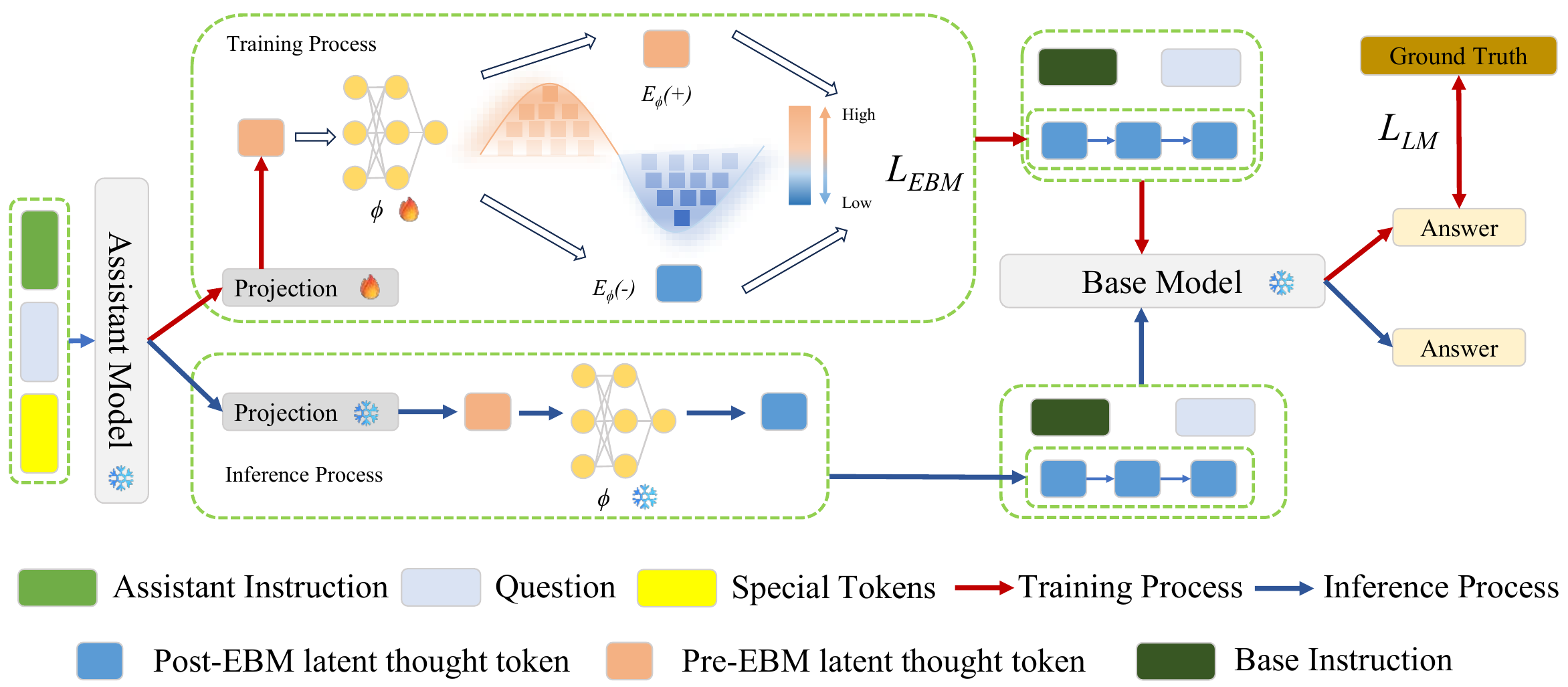}
  \caption{
    Overall architecture of the proposed \method{} framework.
    Given an assistant instruction and a question, along with special tokens, the assistant model first encodes the input and generates a sequence of latent thought tokens that represent intermediate reasoning steps.
    These latent thoughts are then projected through a learnable projection module into the embedding space compatible with the base model.
    An Energy-Based Model (EBM) further refines the projected (Pre-EBM) latent thought tokens via Langevin calibration, assigning lower energy to coherent reasoning states.
    Finally, the calibrated (Post-EBM) latent thought tokens, together with the base instruction and question, are fed into the frozen Base model to generate explicit reasoning steps and produce the final answer.
    This pipeline enables consistent and efficient reasoning from question to answer while preserving the parameters of both the base and assistant models.
    }

  \label{fig:overview1}
\end{figure}

\paragraph{Overview} 
The overall framework of our proposed \method{} is illustrated in Figure~\ref{fig:overview1}.
Unlike previous implicit CoT methods that only rely on continuous representation mapping, our approach introduces an Energy-Based Model (EBM) to explicitly calibrate latent thought tokens within the embedding space. Figure~\ref{fig:overview1} illustrates the overall training and inference processes of the proposed framework.

\paragraph{Training Process} During training, the assistant model first generates latent thought tokens based on the input.
The EBM then calibrates these latent thoughts through contrastive energy learning.
Higher-energy ($E_\phi(+)$) and lower-energy ($E_\phi(-)$) samples are obtained via short-step Langevin dynamics and projected into the thought embedding space, 
where the EBM assigns lower energy to coherent reasoning tokens and higher energy to implausible ones.
The model is optimized using a hinge-style contrastive loss $\mathcal{L}_{\mathrm{h}}(\phi)$ combined with a consistency regularization term. The combined energy-based objective is defined as: $\mathcal{L}_{\mathrm{EBM}} = \mathcal{L}_{h} + \mathcal{L}_{c}$.
This joint objective enables the EBM to learn an energy landscape that effectively distinguishes consistent from inconsistent reasoning trajectories while maintaining smooth latent transitions. The corresponding training procedure is summarized in Algorithm~\ref{alg1}.

\paragraph{Inference Process}
At inference time, the trained EBM performs a lightweight energy-based calibration on the latent thought embeddings generated by the assistant model.
Specifically, a few steps of Langevin updates refine the latent reasoning trajectory by adjusting the embeddings toward lower-energy regions in the thought space.
The calibrated latent thoughts are then fed into the frozen base model for final answer generation.
This framework enables consistent and efficient reasoning without modifying the parameters of either the base or the assistant model. The corresponding inference procedure is summarized in Algorithm~\ref{alg2}.

\subsection{Loss Function Design}
\label{42}

We optimize a combined objective consisting of a language modeling loss $\mathcal{L}_{\mathrm{LM}}$ (computed by the frozen base model given the input) and an energy-based regularization term $\mathcal{L}_{\mathrm{EBM}}$:
\[
    \mathcal{L}_{\mathrm{total}}
    = \mathcal{L}_{\mathrm{LM}}
    + \alpha\,\mathcal{L}_{\mathrm{EBM}},
\]
where $\mathcal{L}_{\mathrm{LM}}$ ensures \textbf{token-level likelihood consistency} 
through standard language modeling, $\mathcal{L}_{\mathrm{EBM}}$ enforces \textbf{latent-level energy consistency} 
by shaping the reasoning dynamics in the continuous latent space, and $\alpha$ balances the two components. These two objectives work in concert to align surface-level linguistic coherence 
with deeper reasoning consistency, resulting in a unified and robust reasoning framework. In our setup, the assistant and base models are frozen; only the energy-based model $\phi$ and the projection are trained.

For the EBM component, we employ a hinge-style contrastive loss between the higher-energy latent $l^{\ell}$ (a negative sample obtained via Langevin sampling) and the lower-energy latent $l^{c}$ (a positive sample from the target data distribution):
\begin{align}
    \mathcal{L}_h
        &= \mathrm{ReLU}\!\left(
              E_\phi(c_t, l^{\ell})
              - E_\phi(c_t, l^{c}) + m
           \right)
\end{align}
Combining with $\mathcal{L}_c= \lambda \|l^{c} - l^{\ell}\|^2$ that calculates the squared L2 distance between \(l^{c}\) and \(l^{\ell }\), 
the overall loss of $\mathcal{L}_{\mathrm{EBM}}$ is 
\begin{align}
\mathcal{L}_{\mathrm{EBM}}
        &= \frac{1}{B}\sum_{b=1}^{B}
           \big(\mathcal{L}_h^{(b)} + \mathcal{L}_c^{(b)}\big)
\end{align}
where $m$ is a margin parameter, $\lambda$ controls the proximity between the positive and negative sample pairs and acts as a regularization term, and $B$ is the batch size.

\subsection{Optimization Process}
\label{43}

To demonstrate the effectiveness of our optimization approach, we provide a corresponding theoretical analysis of the optimization process.


\paragraph{Analysis of $\mathcal{L}_{\mathrm{LM}}$} To examine the gradient flow of our optimization method, we analyze the gradient propagation of the language modeling loss $\mathcal{L}_{\mathrm{LM}}$ through the Langevin sampling steps.
Since each latent state $l^{(s)}$ in the Langevin refinement process depends on the energy function parameter $\phi$, 
the gradient of the language modelling loss $\mathcal{L}_{\mathrm{LM}}$ with respect to $\phi$ can be derived using the chain rule:
\[
    \frac{\partial \mathcal{L}_{\mathrm{LM}}}{\partial \phi}
    = \frac{\partial \mathcal{L}_{\mathrm{LM}}}{\partial l^{(S)}}
      \cdot
      \frac{\partial l^{(S)}}{\partial \phi}
    = \sum_{s=0}^{S-1}
      \frac{\partial \mathcal{L}_{\mathrm{LM}}}{\partial l^{(s+1)}}
      \cdot
      \frac{\partial l^{(s+1)}}{\partial \phi}.
\]
Here, $l^{(S)}$ represents the final latent embedding obtained after $S$ Langevin refinement steps, $l^{(s)}$ denotes the latent thought embedding in the $s$-th iteration ($s = 0, 1, \dots, S-1$), and each update follows the Langevin step
\[
    l^{(s+1)} = l^{(s)} - \eta \nabla_l E_\phi(c_t, l^{(s)}) + \sqrt{2\eta}\,\varepsilon^{(s)},
    \qquad \varepsilon^{(s)} \sim \mathcal{N}(0, I),
\]

where $\varepsilon^{(s)}$ is independent of $\phi$, so $\partial \varepsilon^{(s)}/\partial\phi = 0$. The gradient for a single step \(s+1\) can then be calculated recursively using the chain rule:
\[
    \frac{\partial l^{(s+1)}}{\partial \phi}
    = \frac{\partial l^{(s)}}{\partial \phi}
      - \eta\!\left[
          A^{(s)}\frac{\partial l^{(s)}}{\partial \phi}
          + B^{(s)}
        \right],
\]
where
\[
    A^{(s)} = \frac{\partial}{\partial l}\big(\nabla_l E_\phi(c_t, l^{(s)})\big), B^{(s)} = \frac{\partial(\nabla_l E_\phi(c_t, l^{(s)}))}{\partial \phi},
\]
is the Jacobian of the energy gradient with respect to $l$.
Unrolling this recursion yields the full matrix-product form for the gradient with respect to the final latent state \(l^{(S)}\):
\[
    \frac{\partial \mathcal{L}_{\mathrm{LM}}}{\partial \phi}
    = -\eta \sum_{k=0}^{S-1}
      \frac{\partial \mathcal{L}_{\mathrm{LM}}}{\partial l^{(S)}}
      \left[
          \prod_{j=k+1}^{S-1}(I - \eta A^{(j)})
      \right]
      B^{(k)}.
\]
The detailed derivation is provided in Appendix \ref{theory}.

\paragraph{Analysis of $\mathcal{L}_{\mathrm{EBM}}$} For the EBM loss
\[
    \mathcal{L}_{\mathrm{EBM}}(l^{c}, l^{\ell})
    = \mathrm{ReLU}\!\left(
        E_\phi(c_t, l^{\ell})
        - E_\phi(c_t, l^{c})
        + m
      \right)
      + \lambda \|l^{c} - l^{\ell}\|^2,
\]
its gradient with respect to $\phi$ is
\[
\begin{aligned}
\frac{\partial \mathcal{L}_{\mathrm{EBM}}}{\partial \phi}
&= \mathbf{1}\!\left\{
    E_\phi(c_t, l^{\ell})
    - E_\phi(c_t, l^{c})
    + m > 0
  \right\}
  \left(
    \frac{\partial E_\phi(c_t, l^{\ell})}{\partial \phi}
    - \frac{\partial E_\phi(c_t, l^{c})}{\partial \phi}
  \right) \\
&\quad + 2\lambda\,(l^{c} - l^{\ell})^\top
   \frac{\partial (l^{c} - l^{\ell})}{\partial \phi},
\end{aligned}
\]
where $\mathbf{1}\{\cdot\}$ is the indicator function that activates the hinge loss only when the margin constraint 
$E_\phi(c_t, l^{\ell}) - E_\phi(c_t, l^{c}) + m > 0$ is met (when the positive example does not have a sufficiently lower energy).

\paragraph{Final Gradient Expression} Combining both components, the total gradient of the training objective can be expressed as the sum of the LM-induced gradient propagated through the Langevin chain and the direct EBM gradient:
\[
    \frac{\partial \mathcal{L}_{\mathrm{total}}}{\partial \phi}
    = \frac{\partial \mathcal{L}_{\mathrm{LM}}}{\partial \phi}
      + \alpha\,\frac{\partial \mathcal{L}_{\mathrm{EBM}}}{\partial \phi}.
\]
Writing the LM contribution in shorthand yields
\[
    \frac{\partial \mathcal{L}_{\mathrm{total}}}{\partial \phi}
    = -\eta \sum_{k=0}^{S-1}
      \frac{\partial \mathcal{L}_{\mathrm{LM}}}{\partial l^{(S)}}
      \left[
          \prod_{j=k+1}^{S-1}(I - \eta A^{(j)})
      \right]
     B^{(k)}
      + \alpha\,\frac{\partial \mathcal{L}_{\mathrm{EBM}}}{\partial \phi}.
\]
The gradient formulation above illustrates how the EBM regularizes the latent reasoning trajectory through the Langevin refinement process.
Specifically, the term $\frac{\partial \mathcal{L}_{\mathrm{LM}}}{\partial l^{(S)}}$ propagates the language modeling signal backwards through the sequence of latent updates, 
while the Jacobian product $\prod_{j=k+1}^{S-1}(I - \eta A^{(j)})$ and parameter derivative $B^{(k)}$ 
modulate this signal into adjustments of the energy function $E_\phi$.
As the EBM learns to assign lower energy to consistent reasoning trajectories and higher energy to inconsistent or noisy ones, the Langevin calibration iteratively pushes the latent thought tokens toward globally smoother and more stable configurations.
Consequently, even though the EBM operates only at the latent embedding level, its correction mechanism enforces global coherence across the generated chain of thought and enhances overall consistency in final answers.

\section{Experiments}

To evaluate the effectiveness of our proposed method, we conduct experiments on a diverse set of benchmarks, including datasets of mathematical reasoning, commonsense reasoning, and symbolic reasoning. To ensure a fair comparison, we follow the configurations of the dataset and the model selections adopted in \citet{xu2025softcot}. Detailed descriptions of the baseline methods and datasets are provided in the Appendices \ref{baselines} and \ref{data}. Additional experimental results based on Qwen2.5-7B-Instruct are provided in Appendix~\ref{rd2}.

\subsection{Settings}
\label{settings}

\paragraph{Datasets and Models} 
For the datasets, we have evaluated our method across multiple benchmarks and reasoning paradigms, covering three major categories: mathematical, commonsense, and symbolic reasoning.
For mathematical reasoning, we adopt the GSM8K \citep{cobbe2021training}, ASDiv-aug \citep{miao2021diverse}, and AQuA \citep{ling2017program} datasets; for commonsense reasoning, we use StrategyQA \citep{geva2021did}; and for symbolic reasoning, we employ the Data Understanding (DU) dataset \citep{srivastava2023beyond}. For the models, we have conducted experiments across diverse architectures and parameter scales. The base models include Qwen2.5-7B-Instruct \citep{yang2024qwen2}, Qwen3-8B \citep{yang2025qwen3}, and LLaMA-3.1-8B-Instruct \citep{dubey2024llama}. The assistant models consist of Qwen2.5-7B-Instruct, Qwen2.5-1.5B-Instruct, Qwen2.5-0.5B-Instruct, and Qwen3-0.6B.

\paragraph{Baselines} We compare our approach against three categories of baselines: Zero-Shot, LoRA fine-tuning, and implicit CoT-based reasoning enhancement methods.
For the zero-shot setting, we evaluate Zero-Shot CoT, Zero-Shot Unk, and Zero-Shot Assist-CoT.
For LoRA fine-tuning \citep{hu2022lora}, we apply parameter-efficient fine-tuning using the LoRA framework to assess performance improvements.
For implicit CoT reasoning enhancement, we compare with recent methods such as Coconut \citep{hao2024training} and SoftCoT \citep{xu2025softcot}. Detailed descriptions of these baseline methods are provided in Appendix \ref{baselines}. In addition to the conventional accuracy metric used to evaluate model performance, we introduce a consistency rate metric to quantify the coherence of reasoning across multiple CoT samples. 

\paragraph{Implementation Details} We report the training and inference details of our proposed method. Following the setup in \citet{xu2025softcot} for fair comparison, we jointly train the Energy-Based Model and the projection module using the AdamW optimizer with an initial learning rate of $2\times10^{-5}$. During training, the number of latent thought tokens (num\_thought\_tokens) is set to 32, and we train for 10 epochs with a batch size of 4. For the Langevin calibration process, we apply 3 sampling steps to refine the latent tokens, and set the weighting coefficient $\alpha$ for the energy-based regularization term to 0.1. We also conduct ablation studies to examine the influence of these hyperparameters and other model components. During inference, the main results are obtained using 4 latent thought tokens, and we additionally perform ablation studies on different token numbers to evaluate the trade-off between reasoning depth and efficiency. For the Energy-Based Model, we adopt a multi-layer perceptron architecture as an energy function, following the design choices in \citet{jiang2025learning}.

\subsection{Discussion on LLaMA-3.1-8B-Instruct}
\label{rd1}

Table~\ref{tab:llama3.1-results} reports the reasoning performance of different methods using LLaMA-3.1-8B-Instruct as the base model, LLaMA-3.2-1B-Instruct as the assistant model. With a single reasoning chain (Ours, N=1), the model already achieves an average accuracy of 72.49\%, close to its multi-chain counterpart (Ours, N=10) at 76.60\%.
In contrast, other baselines such as SoftCoT and Coconut exhibit much larger performance gaps between N=1 and N=10, indicating their heavy reliance on multi-sample self-consistency for stable reasoning.
This smaller gap highlights that our energy-based calibration effectively enhances the intrinsic consistency of reasoning within each latent trajectory,
allowing the model to maintain robust performance even with a single inference path.
The improvement is especially evident on ASDiv-Aug and GSM8K, where reducing stochastic variance in multi-step arithmetic reasoning leads to more reliable single-pass results.
Overall, these findings suggest that the proposed latent calibration significantly improves reasoning coherence and stability, reducing dependence on self-consistency aggregation while preserving efficiency.

\begin{table}[t!]
\centering
\caption{Model performance using LLaMA-3.1-8B-Instruct. Here ``N'' denotes the number of sampled reasoning chains. Following prior work, we apply the self-consistency \citep{wang2022self} sampling procedure to aggregate answers across these chains. Notably, our method achieves strong performance even with a single reasoning chain ($N=1$), demonstrating high inherent consistency without requiring multi-sample aggregation.}
\resizebox{\linewidth}{!}{
\begin{tabular}{lcccccc}
\toprule
\textbf{Model} & \textbf{GSM8K} & \textbf{ASDiv-Aug} & \textbf{AQuA} & \textbf{StrategyQA} & \textbf{DU} & \textbf{Avg.} \\
 & & \textbf{Mathematical} & & \textbf{Commonsense} & \textbf{Symbolic} & \\
\midrule
Zero-Shot CoT & $79.61_{\pm0.81}$ & $86.78_{\pm0.63}$ & $54.65_{\pm2.43}$ & $65.63_{\pm3.31}$ & $54.40_{\pm2.40}$ & $68.21$ \\
Zero-Shot CoT-Unk & $79.95_{\pm0.59}$ & $86.90_{\pm0.41}$ & $55.28_{\pm1.88}$ & $66.16_{\pm2.70}$ & $54.16_{\pm1.46}$ & $68.49$ \\
Zero-Shot Assist-CoT & $80.76_{\pm1.53}$ & $86.96_{\pm0.46}$ & $55.83_{\pm2.98}$ & $66.55_{\pm3.99}$ & $58.24_{\pm3.56}$ & $69.67$ \\
LoRA Fine-Tuning & $75.66_{\pm0.00}$ & $86.67_{\pm0.00}$ & $52.36_{\pm0.00}$ & - & - & - \\
Coconut & $76.12_{\pm0.00}$ & $86.80_{\pm0.00}$ & $53.15_{\pm0.00}$ & - & - & - \\
SoftCoT & $81.03_{\pm0.42}$ & $87.19_{\pm0.40}$ & $56.30_{\pm1.67}$ & $69.04_{\pm1.23}$ & $59.04_{\pm1.93}$ & $70.52$ \\
Ours & $\textbf{85.26$_{\pm0.04}$}$ & $\textbf{88.20$_{\pm0.08}$}$ & $\textbf{58.14$_{\pm0.18}$}$ & $\textbf{69.54$_{\pm0.33}$}$ & $\textbf{61.32$_{\pm0.26}$}$ & $\textbf{72.49}$ \\
\midrule
Zero-Shot CoT ($N=10$) & $90.36_{\pm0.40}$ & $89.23_{\pm0.17}$ & $63.23_{\pm0.86}$ & $70.13_{\pm0.47}$ & $65.76_{\pm1.54}$ & $68.21$ \\

Zero-Shot Assist-CoT ($N=10$) & $90.43_{\pm0.69}$ & $89.48_{\pm0.36}$ & $63.62_{\pm0.99}$ & $70.48_{\pm0.68}$ & $65.84_{\pm1.93}$ & $69.67$ \\
Coconut ($N=10$)& $87.03_{\pm0.00}$ & $88.44_{\pm0.00}$ & $61.81_{\pm0.00}$ & - & - & - \\
SoftCoT ($N=10$) & $90.63_{\pm0.39}$ & $89.75_{\pm0.29}$ & $65.51_{\pm0.72}$ & $71.14_{\pm0.10}$ & $67.36_{\pm1.12}$ & $76.88$ \\

Ours ($N=10$) & $90.48_{\pm0.08}$ & $89.33_{\pm0.05}$ & $65.45_{\pm0.32}$ & $71.18_{\pm0.20}$ & $67.93_{\pm0.25}$ & $76.87$ \\
\bottomrule
\end{tabular}}

\label{tab:llama3.1-results}
\end{table}

\subsection{Discussion on Qwen3-8B}
\label{rd3}

Table~\ref{tab:qwen3-results} reports the results of five reasoning benchmarks using Qwen3-8B as the base model, Qwen3-0.6B as the assistant model.
Overall, our method continues to outperform all baselines across different reasoning types, 
demonstrating that the proposed energy-based calibration generalizes effectively to larger-scale models.
Compared to SoftCoT, our approach achieves an average improvement of +2.1\%, confirming that refining latent thought tokens with the learned energy function 
consistently improves reasoning stability and coherence. The self-consistency variant (Ours (N=10)) further increases the average accuracy to 85.95\%, 
surpassing all other multi-sample baselines. 
This improvement highlights that energy-based calibration and self-consistency sampling are complementary:
The EBM reduces local inconsistency in latent reasoning trajectories, while multiple sampling chains provide a more reliable aggregation of reasoning outcomes.
 Performance gains are most pronounced on symbolic (DU) and commonsense (StrategyQA) tasks,
indicating that the proposed latent calibration effectively mitigates reasoning variance even in abstract or open-domain settings.

\begin{table}[t!]
\centering
\caption{Model performance using Qwen3-8B.}
\resizebox{\linewidth}{!}{
\begin{tabular}{lcccccc}
\toprule
\textbf{Model} & \textbf{GSM8K} & \textbf{ASDiv-Aug} & \textbf{AQuA} & \textbf{StrategyQA} & \textbf{DU} & \textbf{Avg.} \\
 & & \textbf{Mathematical} & & \textbf{Commonsense} & \textbf{Symbolic} & \\
\midrule
Zero-Shot CoT & $91.86_{\pm0.41}$ & $91.70_{\pm0.26}$ & $70.00_{\pm1.56}$ & $69.87_{\pm0.35}$ & $80.32_{\pm2.32}$ & $80.75$ \\
Zero-Shot Assist-CoT & $91.90_{\pm0.50}$ & $91.64_{\pm0.16}$ & $70.16_{\pm2.35}$ & $69.78_{\pm0.33}$ & $80.56_{\pm1.15}$ & $80.81$ \\
Coconut & $87.95_{\pm0.00}$ & $89.40_{\pm0.00}$ & $68.50_{\pm0.00}$ & - & - & - \\
SoftCoT & $92.48_{\pm0.29}$ & $91.83_{\pm0.19}$ & $75.04_{\pm2.68}$ & $70.17_{\pm0.63}$ & $81.21_{\pm0.70}$ & $83.82$ \\
Ours & $\textbf{92.82$_{\pm0.09}$}$ & $\textbf{92.22$_{\pm0.18}$}$ & $\textbf{86.02$_{\pm0.57}$}$ & $\textbf{70.96$_{\pm0.36}$}$ & $\textbf{82.29$_{\pm0.87}$}$ & $\textbf{84.86}$ \\
\midrule
Zero-Shot CoT ($N=10$) & $92.22_{\pm0.47}$ & $91.97_{\pm0.13}$ & $76.77_{\pm0.62}$ & $70.96_{\pm0.15}$ & $84.56_{\pm0.61}$ & $83.30$ \\

Zero-Shot Assist-CoT ($N=10$) & $92.68_{\pm0.17}$ & $91.91_{\pm0.28}$ & $76.77_{\pm0.79}$ & $70.92_{\pm0.28}$ & $84.80_{\pm1.17}$ & $83.82$ \\

Coconut ($N=10$) & $90.37_{\pm0.00}$ & $90.37_{\pm0.00}$ & $76.38_{\pm0.00}$ & - & - & - \\

SoftCoT ($N=10$) & $93.19_{\pm0.32}$ & $92.14_{\pm0.15}$ & $80.63_{\pm1.90}$ & $71.18_{\pm0.15}$ & $84.01_{\pm0.22}$ & $84.87$ \\

Ours ($N=10$) & $93.28_{\pm0.25}$ & $92.26_{\pm0.04}$ & $88.19_{\pm0.58}$ & $71.11_{\pm0.37}$ & $84.89_{\pm0.29}$ & $85.95$ \\
\bottomrule
\end{tabular}}
\label{tab:qwen3-results}
\end{table}

\subsection{Ablation Study}
\label{rd3}

\begin{figure}[t!]
  \setlength{\abovecaptionskip}{-0.1cm}
  \setlength{\belowcaptionskip}{-0.2cm}
  \vspace{-.1cm}
  \centering

    \includegraphics[width=0.9\columnwidth]{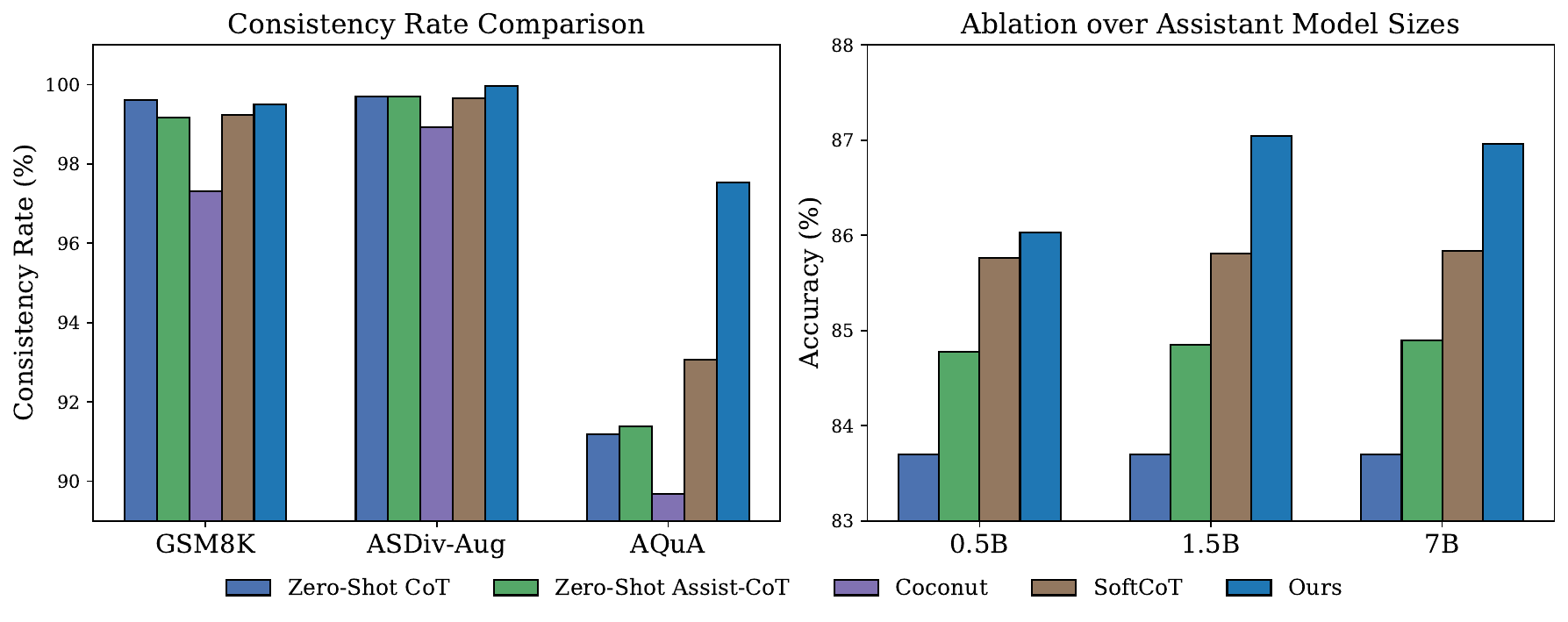}
  \caption{\textbf{Left Figure:} Consistency rate comparison on GSM8K, ASDiv-Aug, and AQuA using Qwen3-8B as the base model and Qwen3-0.6B as the assistant model. \textbf{Right Figure:} Ablation over assistant model sizes on GSM8K using Qwen2.5-7B-Instruct as the base model.}
  \label{fig:ablation}

\end{figure}

Figure~\ref{fig:ablation} presents an ablation study that evaluates the consistency and scalability of the proposed energy-based calibration framework.
On the left, we assess the consistency rate using Qwen3-8B as the base model and Qwen3-0.6B as the assistant model. 
Our approach achieves nearly perfect consistency across all datasets, surpassing SoftCoT and other baselines by a large margin. 
This confirms that correcting latent thought tokens through the energy model effectively mitigates stochastic deviations in implicit reasoning, leading to more stable multi-step reasoning outcomes. On the right, we conduct an ablation over different assistant model sizes using Qwen2.5-7B-Instruct on GSM8K. 
Performance steadily improves as the assistant model grows from 0.5B to 7B parameters, reflecting that stronger assistants provide more informative latent priors. 
Nevertheless, our method consistently achieves the best results regardless of model scale, demonstrating that the proposed calibration is model-agnostic and robust even with lightweight assistants. 
These results highlight that our approach improves reasoning consistency while maintaining scalability across heterogeneous model configurations.

\begin{figure}[t!]
  \setlength{\abovecaptionskip}{-0.1cm}
  \setlength{\belowcaptionskip}{-0.2cm}
  \vspace{-.1cm}
  \centering

    \includegraphics[width=1\columnwidth]{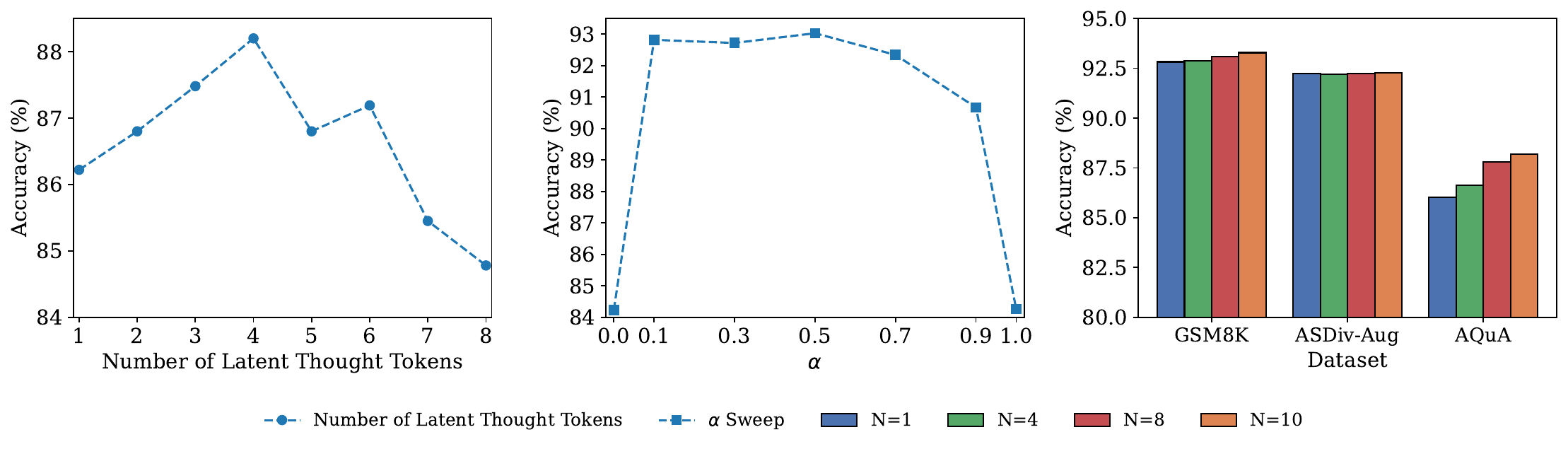}
  \caption{\textbf{Left Figure:} Results on ASDiv-Aug using Qwen2.5-7B-Instruct as the base model and Qwen2.5-1.5B-Instruct as the assistant model, with varying numbers of latent thought tokens. \textbf{Middle Figure:} Results on GSM8K using Qwen3-8B as the base model and Qwen3-0.6B as the assistant model, where $\alpha$ controls the relative strength of the energy-based regularization term $\mathcal{L}_{\mathrm{EBM}}$. \textbf{Right Figure:} Comparison across different numbers of reasoning chains $N$ (pass@$N$) using Qwen3-8B as the base model. While larger $N$ slightly improves performance due to answer aggregation under the self-consistency setting, our model already achieves strong single-chain accuracy ($N=1$), highlighting that energy-based calibration substantially enhances reasoning consistency without relying on multi-chain sampling. }
  \vspace{-.1cm}
  \label{fig:ablation1}

\end{figure}

Figure~\ref{fig:ablation1} presents the sensitivity analysis of our proposed framework with respect to the number of latent thought tokens, 
the EBM weighting coefficient $\alpha$, and the number of reasoning chains $N$. 
On the left, we observe that increasing the number of latent thought tokens initially improves accuracy, 
as more latent steps allow the model to capture finer-grained reasoning patterns in the embedding space. 
However, performance saturates and then declines beyond four tokens, implying that excessive latent reasoning introduces unstable optimization, 
which can distort the energy landscape and hinder convergence. 
In the middle, we examine the influence of $\alpha$, which balances the language modeling loss $\mathcal{L}_{\mathrm{LM}}$ and the energy-based term $\mathcal{L}_{\mathrm{EBM}}$. 
A moderate value of $\alpha$ (around 0.5) yields the highest accuracy, indicating that partial energy regularization is sufficient to guide consistent reasoning 
without overpowering the linguistic prior of the base model. 
When $\alpha$ approaches 0 or 1, performance drops sharply — either due to insufficient calibration or excessive energy dominance. 
On the right, we compare the effect of different numbers of reasoning chains $N$ (pass@$N$) under the self-consistency setting. 
Although accuracy improves slightly with larger $N$, our method already achieves competitive results at $N=1$, 
demonstrating that the proposed energy-based calibration substantially enhances single-chain consistency, 
reducing the need for costly multi-chain aggregation. 
These findings collectively highlight that both the number of latent thought tokens and the strength of energy calibration 
play critical roles in shaping stable and efficient reasoning trajectories, while the model’s high single-chain accuracy reflects 
its intrinsic reasoning consistency.

\section{Discussion}

This work introduces an energy-based calibration framework that bridges the gap between explicit Chain-of-Thought (CoT) reasoning and continuous latent thought token optimization. 
Our experiments demonstrate that our EBM-CoT consistently improves reasoning accuracy and stability across mathematical, commonsense, and symbolic reasoning benchmarks while maintaining high computational efficiency.
Despite these promising results, several limitations remain. 
First, the computational overhead introduced by the Langevin refinement, though lightweight in our setting ($S=3$ steps), may still scale unfavorably for extremely large models or long reasoning sequences. 
Second, the current energy function is implemented as a shallow MLP, which may limit its expressiveness in modelling complex latent dependencies. 
Future work will explore more structured or hierarchical energy formulations and adaptive Langevin updates to improve scalability and dynamic reasoning control. Overall, this study highlights the potential of energy-based modelling as a powerful and general mechanism for reasoning calibration in LLMs.
We hope this work inspires further exploration of the implicit CoT field.

\bibliography{iclr2026_conference}
\bibliographystyle{iclr2026_conference}

\appendix

\section{Theoretical Properties}
\label{theory}

\subsection{Residual Energy-based Conditional Model}

Given a condition $c_t = (x, l_{<t})$, we define the conditional energy-based model (EBM) over latent thought embeddings $l_t$ as
\begin{align}
    \tilde{p}_\phi(l_t \mid c_t)
    &= \frac{1}{Z_\phi(c_t)}\, p_{\mathrm a}(l_t \mid c_t)\,
       \exp\!\left(-E_\phi(c_t, l_t)/{T}\right),
\end{align}
where $p_{\mathrm a}(l_t \mid c_t)$ denotes the assistant-provided latent thought token distribution, $E_\phi(c_t, l_t)$ is the energy function parameterized by $\phi$, $T$ is a temperature, and $Z_\phi(c_t)$ is the normalization constant (partition function):
\begin{align}
Z_\phi(c_t)
&= \mathbb{E}_{v\sim p_{\mathrm a}(\cdot\mid c_t)}\!\left[\exp\!\left(-\frac{E_\phi(c_t,v)}{T}\right)\right]
= \int \exp\!\left(-\frac{E_\phi(c_t,v)}{T}\right)p_{\mathrm a}(v\mid c_t)\,dv.
\end{align}
The normalized form is therefore
\[
\tilde{p}_\phi(l_t\mid c_t)
= \frac{
        p_{\mathrm a}(l_t \mid c_t)\,
        \exp\!\left(-E_\phi(c_t, l_t)/{T}\right)
       }{
        \mathbb{E}_{v\sim p_{\mathrm a}(\cdot\mid c_t)}
        \!\left[\exp\!\left(-E_\phi(c_t, v)/{T}\right)\right]}.
\]

\subsection{Chain-of-Thought Factorization}

For a latent sequence $L = \{l_1, \dots, l_T\}$ conditioned on input $x$, we assume the auto-regressive factorization
\[
    p_{\mathrm a}(L \mid x) = \prod_{t=1}^T p_{\mathrm a}(l_t \mid c_t),
\]
where $c_t = (x, l_{<t})$ denotes the context up to step $t$.
Replacing the assistant prior with the energy-calibrated latent distribution yields
\[
    \tilde{p}_\phi(L \mid x)
    = \prod_{t=1}^T \tilde{p}_\phi(l_t \mid c_t),
\]
which is the distribution our latent EBM aims to approximate or induce.

\subsection{The relationship between $l^{(S)}$ and $l^{(s)}$}

The relationship between $l^{(S)}$ and $\{l^{(s)}\}_{s=0}^{S-1}$ can be expressed recursively as
\[
    l^{(s+1)} = l^{(s)} - \eta \nabla_l E_\phi(c_t, l^{(s)}) + \sqrt{2\eta}\,\varepsilon^{(s)},
    \qquad \varepsilon^{(s)} \sim \mathcal{N}(0, I),
\]
and by unrolling all $S$ updates, the final latent state can be written as
\[
    l^{(S)} = l^{(0)} - \eta \sum_{s=0}^{S-1} \nabla_l E_\phi(c_t, l^{(s)}) + \sqrt{2\eta}\sum_{s=0}^{S-1}\varepsilon^{(s)}.
\]
Therefore, $l^{(S)}$ is an accumulated result of all intermediate latent states $\{l^{(s)}\}$,
each contributing a gradient-based correction and a stochastic perturbation toward a lower-energy (i.e., more consistent) region of the latent space.

\subsection{Langevin Calibration Process}

To draw samples from the energy-based latent model, we apply overdamped Langevin dynamics directly in the continuous latent (embedding) space.
At iteration $s$, the candidate latent state $l^{(s)}$ evolves according to the discretized update
\[
    l^{(s+1)} = l^{(s)}
      - \eta \nabla_l E_\phi(c_t, l^{(s)})
      + \sqrt{2\eta}\,\varepsilon^{(s)},\qquad
      \varepsilon^{(s)} \sim \mathcal{N}(0, I),
\]
where $\eta$ is the step size and the noise scale $\sqrt{2\eta}$ ensures consistency with the continuous Langevin Stochastic Differential Equation (SDE).
This update defines a conditional Markov chain
\[
    l^{(s+1)} \sim p_\phi(l^{(s+1)} \mid l^{(s)}, c_t),
\]
which, under standard regularity and small-step assumptions, targets the stationary conditional distribution
\[
    p_\phi(l \mid c_t) \propto p_{\mathrm a}(l\mid c_t)\exp\!\big(-E_\phi(c_t,l)/T\big),
\]
up to the choice of initialization and proposal.

In the conditional setting, $c_t = (x, l_{<t})$ is a fixed context comprising the input $x$ and the previously generated latent thoughts, while $l^{(s)}$ denotes the current latent candidate in the inner-loop sampling.
Each update $l^{(s+1)}$ depends only on $l^{(s)}$, forming a Markov process conditioned on $c_t$.
In practice, we initialize $l^{(0)}$ with the assistant output $l^+ \sim p_{\mathrm a}(\cdot \mid c_t)$,
which typically lies in a higher-energy region of the latent space.
After $S$ Langevin refinement steps, we obtain the calibrated sample $l^{c} \equiv l^{(S)}$,
corresponding to a lower-energy and more consistent latent thought,
while $l^{\ell} \equiv l^+$ denotes the unrefined, higher-energy latent sample provided by the assistant model.

\subsection{Gradient propagation through Langevin steps.}
The Langevin update is
\[
l^{(s+1)} = l^{(s)} - \eta \nabla_l E_\phi(c_t, l^{(s)}) + \sqrt{2\eta}\,\varepsilon^{(s)},\qquad
\varepsilon^{(s)}\sim\mathcal{N}(0,I).
\]
Since the final loss $\mathcal{L}_{\mathrm{LM}}$ depends on $l^{(S)}$, by the chain rule
\[
\frac{\partial \mathcal{L}_{\mathrm{LM}}}{\partial \phi}
= \frac{\partial \mathcal{L}_{\mathrm{LM}}}{\partial l^{(S)}}\cdot\frac{\partial l^{(S)}}{\partial\phi}.
\]
Let $g^{(s)}=\nabla_l E_\phi(c_t,l^{(s)})$ and define
\[
A^{(s)}=\frac{\partial g^{(s)}}{\partial l},\qquad
C^{(s)}=\frac{\partial g^{(s)}}{\partial\phi}.
\]
Then each step satisfies the recursion
\[
\frac{\partial l^{(s+1)}}{\partial\phi}
= (I-\eta A^{(s)})\frac{\partial l^{(s)}}{\partial\phi} - \eta C^{(s)},
\qquad
\frac{\partial l^{(0)}}{\partial\phi}=0.
\]
Unrolling the recursion gives
\[
\frac{\partial l^{(S)}}{\partial\phi}
= -\eta\sum_{k=0}^{S-1}
  \Big(\prod_{j=k+1}^{S-1}(I-\eta A^{(j)})\Big) C^{(k)},
\]
and hence
\[
\frac{\partial \mathcal{L}_{\mathrm{LM}}}{\partial \phi}
= -\eta \sum_{k=0}^{S-1}
  \left(\frac{\partial \mathcal{L}_{\mathrm{LM}}}{\partial l^{(S)}}
        \prod_{j=k+1}^{S-1}(I-\eta A^{(j)})\right) C^{(k)}.
\;
\]

\section{Algorithm Description}
\label{alg}

\textbf{Training Process.}
During training, the assistant model first generates latent thought tokens from the input,
which are projected into the base model's embedding space via a learnable projection network $P$.
The Energy-Based Model (EBM) $E_\phi$ evaluates these latent tokens by assigning lower energy to task-relevant and consistent thoughts
and higher energy to incoherent ones.
To refine the latent representations, we perform $S$ Langevin sampling steps with step size $\eta$,
iteratively updating the latent variable $l^c$ using the gradient of the energy function and Gaussian noise.
The refined latent embeddings are then injected into the base model to compute the language modelling loss $\mathcal{L}_{\mathrm{LM}}$.
The EBM parameters $\phi$ and projection $P$ are optimized jointly by minimizing a weighted objective
$\mathcal{L} = \mathcal{L}_{\mathrm{LM}} + \alpha \mathcal{L}_{\mathrm{EBM}}$,
where $\alpha$ controls the strength of the energy-based regularization.
Following our experimental setup, we use $S=3$ Langevin steps and set $\alpha=0.1$ to balance both objectives.

\begin{algorithm}[H]
\caption{Training process with EBM Correction}
\label{alg1}
\begin{algorithmic}[1]
\REQUIRE Assistant model $p_a$, projection $P$, energy network $E_\phi$, step size $\eta$, steps $S$, loss weight $\alpha$
\FOR{each batch $(x, y)$}
    \STATE Compute base embeddings $X \leftarrow \text{embed\_base}(x)$
    \STATE Generate assistant hidden states $H_a \leftarrow p_a(x)$
    \STATE Project to base dimension $T \leftarrow P(H_a)$
    \STATE Extract latent thought $l^{\ell}$ from $T$
    \STATE Initialize corrected latent $l^{c} \leftarrow l^{\ell}$
    \FOR{$s = 1$ to $S$} 
    
        \STATE Compute gradient $g \leftarrow \nabla_{l} E_\phi(c_t, l^{c})$
        \STATE Sample noise $\xi_s \sim \mathcal{N}(0, I)$
        \STATE Update latent $l^{c} \leftarrow l^{c} - \eta g + \sqrt{2\eta}\,\xi_s$
    \ENDFOR
    \STATE Replace latent thought tokens in $X$ with $l^{c}$ to get $X^{\mathrm{mod}}$
    \STATE Forward base model to obtain $\mathcal{L}_{\mathrm{LM}}$
    \STATE Compute hinge loss $\mathcal{L}_h = \mathrm{ReLU}(E_\phi(l^{\ell}) - E_\phi(l^{c}) + m)$
    \STATE Compute consistency loss $\mathcal{L}_c = \lambda \|l^{c}-l^{\ell}\|^2$
    \STATE $\mathcal{L}_{\mathrm{EBM}} \leftarrow \mathcal{L}_h + \mathcal{L}_c$
    \STATE Total loss $\mathcal{L} = \mathcal{L}_{\mathrm{LM}} + \alpha \mathcal{L}_{\mathrm{EBM}}$
    \STATE Backpropagate and update $\phi$ and  $P$
\ENDFOR
\end{algorithmic}
\end{algorithm}

\textbf{Inference Process.}
At inference time, the assistant model provides the initial latent thoughts $l^{\ell}$,
which are refined through several deterministic Langevin updates (without noise) to obtain the calibrated latents $l^{c}$.
The updated embeddings are then fed into the frozen base model for final output generation.
This process enables consistent reasoning and robust predictions without altering the parameters of either the base or assistant model.

\begin{algorithm}[H]
\caption{Inference process with EBM Correction}
\label{alg2}
\begin{algorithmic}[1]
\STATE Given input $x$, compute base embeddings $X$
\STATE Assistant forward $\rightarrow$ latent $l^{\ell}$
\STATE Initialize $l^{c} \leftarrow l^{\ell}$
\FOR{$s = 1$ to $S$}
    \STATE $g \leftarrow \nabla_{l} E_\phi(c_t, l^{c})$
    \STATE $l^{c} \leftarrow l^{c} - \eta g$ \COMMENT{(noise optional during inference)}
\ENDFOR
\STATE Replace latent thought tokens in $X$ with $l^{c}$
\end{algorithmic}
\end{algorithm}

\section{Baselines}
\label{baselines}

We compare our approach against three major categories of baselines:
(\textit{i}) zero-shot prompting methods,
(\textit{ii}) parameter-efficient fine-tuning via LoRA,
and (\textit{iii}) implicit chain-of-thought (CoT) reasoning enhancement approaches.
This setup allows us to assess the contribution of directional alignment under different reasoning paradigms and adaptation strategies.

\paragraph{Zero-shot prompting baselines.}
We first evaluate several strong zero-shot prompting methods that require no additional fine-tuning:
\begin{itemize}
    \item \textbf{Zero-Shot CoT} uses the simple prompt ``\textit{Let's think step by step}'' to elicit multi-step reasoning without any labeled demonstrations.
    This serves as a standard baseline for evaluating the intrinsic reasoning capability of large language models.
    \item \textbf{Zero-Shot Unk} extends Zero-Shot CoT by introducing uncertainty indicators (e.g., ``\textit{I'm not sure, but let's reason carefully}'') to encourage more cautious and calibrated reasoning.
    It provides insight into how uncertainty modulation affects reasoning reliability.
    \item \textbf{Zero-Shot Assist-CoT} supplies auxiliary reasoning traces or scaffolds generated by another model or previous step, assisting the target model in refining its reasoning path.
    This simulates collaborative or staged reasoning without additional training.
\end{itemize}

To further improve the robustness of zero-shot reasoning, we also include the \textbf{Self-Consistency (SC)} baseline~\citep{wang2022self}.
SC samples multiple diverse reasoning paths (we set $N = 10$) using temperature-based decoding, and selects the most consistent final answer by majority voting.
This approach has been shown to significantly improve reasoning stability and reduce stochasticity in CoT inference.

\paragraph{LoRA fine-tuning baselines.}
To compare our approach with parameter-efficient adaptation methods, we employ \textbf{LoRA}~\citep{hu2022lora} fine-tuning.
LoRA introduces low-rank trainable adapters into existing linear layers while freezing the majority of model parameters.
We fine-tune LoRA adapters on each task-specific dataset using a small learning rate and evaluate the resulting models on the corresponding test splits.
This baseline measures the benefits of explicit fine-tuning relative to our merging-based adaptation, which does not require retraining.

\paragraph{Implicit CoT-based reasoning enhancement.}
Finally, we compare with recent implicit reasoning methods that enhance the internal reasoning ability of language models without explicit step supervision:
\begin{itemize}
    \item \textbf{Coconut}~\citep{hao2024training} learns latent reasoning structures by aligning intermediate reasoning representations with the final answer, encouraging consistency between thought processes and outputs.
    \item \textbf{SoftCoT}~\citep{xu2025softcot} distills soft reasoning distributions from a teacher model into the student, enabling smoother reasoning trajectories and better generalization across reasoning tasks.
\end{itemize}

Together, these baselines cover three complementary paradigms:
prompt-based reasoning (\textit{Zero-Shot CoT, Self-Consistency}),
parameter-efficient fine-tuning (\textit{LoRA}),
and implicit reasoning enhancement (\textit{Coconut, SoftCoT}).
Our experiments aim to demonstrate that directional alignment provides consistent gains across all these settings while maintaining training-free efficiency.

\section{Datasets}
\label{data}

We evaluate our proposed method on five publicly available reasoning datasets covering
mathematical, commonsense, and symbolic reasoning.

\paragraph{Mathematical reasoning.}
We use three mathematical reasoning benchmarks: GSM8K~\citep{cobbe2021training},
ASDiv-Aug~\citep{miao2021diverse}, and AQuA~\citep{ling2017program}.
These datasets require arithmetic and algebraic reasoning with varying complexity levels.
GSM8K contains 7473 training and 1319 evaluation samples,
where each question demands a free-form numerical answer.
ASDiv-Aug extends the ASDiv dataset with data augmentation,
providing 4183 training and 1038 test samples,
also with numerical answers.
AQuA includes 97467 training and 254 evaluation questions
in a multiple-choice format.
Together, these datasets evaluate a model’s ability to perform step-by-step quantitative reasoning.

\paragraph{Commonsense reasoning.}
We evaluate on StrategyQA~\citep{geva2021did},
a dataset designed to test implicit multi-hop commonsense reasoning.
Each question requires combining multiple unspoken facts to produce a binary yes/no answer.
It consists of 1832 training and 458 evaluation samples,
challenging models to perform compositional and non-explicit reasoning.

\paragraph{Symbolic reasoning.}
Finally, we adopt the Date Understanding (DU) task
from BIG-Bench~\citep{srivastava2023beyond}.
DU evaluates a model’s ability to resolve temporal relationships expressed in natural language.
It is a purely evaluation-only benchmark with 369 samples and no training split available.
Each question provides several options, requiring symbolic comparison and logical deduction.

\section{Discussion on Qwen2.5-7B-Instruct}
\label{rd2}

Table~\ref{tab:qwen2.5-results} summarizes the model performance on five reasoning benchmarks using Qwen2.5-7B-Instruct as base model, Qwen2.5-1.5B-Instruct as assistant model.
Our proposed method consistently outperforms all baselines across diverse reasoning domains, including mathematical (GSM8K, ASDiv-A) and commonsense (StrategyQA) tasks.
Compared with SoftCoT, which relies solely on continuous representation mapping, our energy-based calibration achieves an average improvement of +3.8\%, demonstrating that explicitly refining the latent thought tokens leads to more stable and coherent reasoning trajectories.

The self-consistency variant (Ours (N=10)) further amplifies this gain, achieving the highest overall average accuracy of 82.73\%.
By sampling multiple calibrated reasoning paths and aggregating their outcomes, the model effectively reduces stochastic variance in the latent reasoning space.
This indicates that the proposed energy-based refinement complements self-consistency sampling by aligning latent trajectories toward lower-energy, high-consistency regions.
In contrast, prior methods such as SoftCoT or LoRA fine-tuning lack such global regularization, leading to greater variation in reasoning quality across sampled chains.

Notably, the largest gains appear on symbolic and commonsense datasets (DU, StrategyQA), suggesting that our calibration mechanism generalizes beyond arithmetic reasoning and enhances structural consistency in broader reasoning contexts.
Overall, these results validate the effectiveness of our approach in improving both the accuracy and stability of multi-step reasoning in large language models.

\begin{table}[t!]
\centering
\caption{Model performance using Qwen2.5-7B-Instruct.
}
\resizebox{\linewidth}{!}{
\begin{tabular}{lcccccc}
\toprule
\textbf{Model} & \textbf{GSM8K} & \textbf{ASDiv-Aug} & \textbf{AQuA} & \textbf{StrategyQA} & \textbf{DU} & \textbf{Avg.} \\
 & & \textbf{Mathematical} & & \textbf{Commonsense} & \textbf{Symbolic} & \\
\midrule
Zero-Shot CoT & $83.70_{\pm0.78}$ & $87.19_{\pm0.28}$ & $64.53_{\pm3.27}$ & $49.65_{\pm3.18}$ & $66.40_{\pm2.26}$ & $70.29$ \\
Zero-Shot CoT-Unk & $84.12_{\pm0.71}$ & $86.94_{\pm0.89}$ & $64.72_{\pm2.06}$ & $50.74_{\pm1.90}$ & $66.48_{\pm1.43}$ & $70.60$ \\
Zero-Shot Assist-CoT & $84.85_{\pm0.35}$ & $88.63_{\pm1.05}$ & $64.96_{\pm2.83}$ & $52.71_{\pm2.65}$ & $67.04_{\pm2.84}$ & $71.64$ \\
\midrule
LoRA Fine-Tuning & $81.80_{\pm0.00}$ & $86.80_{\pm0.00}$ & $62.60_{\pm0.00}$ & - & - & - \\
Coconut & $82.49_{\pm0.00}$ & $86.90_{\pm0.00}$ & $63.39_{\pm0.00}$ & - & - & - \\
SoftCoT & $85.81_{\pm1.82}$ & $88.90_{\pm1.01}$ & $72.44_{\pm2.19}$ & $60.61_{\pm1.55}$ & $67.52_{\pm2.92}$ & $75.06$ \\
Ours & $\textbf{87.04$_{\pm0.23}$}$ & $\textbf{90.70$_{\pm0.24}$}$ & $\textbf{75.59$_{\pm0.96}$}$ & $\textbf{67.47$_{\pm0.22}$}$ & $\textbf{73.37$_{\pm1.02}$}$ & $\textbf{78.83}$ \\
Ours ($N=10$) & $92.75_{\pm0.04}$ & $91.56_{\pm0.09}$ & $83.37_{\pm0.59}$ & $69.28_{\pm0.10}$ & $76.69_{\pm0.39}$ & $82.73$ \\
\bottomrule
\end{tabular}}

\label{tab:qwen2.5-results}
\end{table}

\section{Devices}

In the experiments, we conduct all methods on a local Linux server equipped with one AMD EPYC 7742 64-Core Processor (128 logical threads). 
All methods are implemented using the PyTorch framework, and all models are trained on NVIDIA A100-SXM4-80GB GPUs (80GB HBM2 memory).

\end{document}